\documentclass[10pt,twocolumn,letterpaper]{article}

\usepackage[pagenumbers]{cvpr} 

\usepackage[dvipsnames]{xcolor}

\usepackage{multirow}
\usepackage{xcolor}
\usepackage{lipsum}
\usepackage{amsmath}

\definecolor{cvprblue}{rgb}{0.21,0.49,0.74}
\usepackage[pagebackref,breaklinks,colorlinks,citecolor=cvprblue]{hyperref}

\def\paperID{16026} 
\def\confName{CVPR}
\def\confYear{2024}

\title{Windowed-FourierMixer: Enhancing Clutter-Free Room Modeling with Fourier Transform}

\author{Bruno Henriques, Benjamin Allaert and Jean-Philippe Vandeborre\\
IMT Nord Europe, Institut Mines Télécom, Univ. Lille Centre for Digital Systems, F-59000 Lille, France \\
{\tt\small \{f\_author.s\_author\}@imt-nord-europe.fr}}

\begin{document}
\maketitle

\begin{abstract}
With the growing demand for immersive digital applications, the need to understand and reconstruct 3D scenes has significantly increased. In this context, inpainting indoor environments from a single image plays a crucial role in modeling the internal structure of interior spaces as it enables the creation of textured and clutter-free reconstructions.
While recent methods have shown significant progress in room modeling, they rely on constraining layout estimators to guide the reconstruction process. These methods are highly dependent on the performance of the structure estimator and its generative ability in heavily occluded environments.
In response to these issues, we propose an innovative approach based on a U-Former architecture and a new Windowed-FourierMixer block, resulting in a unified, single-phase network capable of effectively handle human-made periodic structures such as indoor spaces. This new architecture proves advantageous for tasks involving indoor scenes where symmetry is prevalent, allowing the model to effectively capture features such as horizon/ceiling height lines and cuboid-shaped rooms.
Experiments show the proposed approach outperforms current state-of-the-art methods on the Structured3D~\cite{zheng2020structured3d} dataset demonstrating superior performance in both quantitative metrics and qualitative results. Code and models will be made publicly available.
\end{abstract}

\section{Introduction}
\label{sec:intro}
    \paragraph{}
    The ability to understand and reconstruct 3d environments is of paramount importance in immersive digital applications. In the context of scene reconstruction for immersive digital applications, the separation of structural (layout and geometry) and contextual (object-related) information is highly advantageous. This separation allows users to augment and diminish their surroundings and offers developers finer-grained control over the virtually reconstructed environment. However, the process of separating both types of information can pose challenges, especially when transitioning from cluttered to clutter-free 3D environments. Removing unwanted elements and obstructions during this separation can lead to gaps and missing regions, which disrupt the overall visual coherence of the reconstructed scene.
    
    Image inpainting plays a vital role in ensuring that the reconstructed 3D environment remains visually coherent. Image inpainting fills in missing or damaged areas of an image, making it particularly useful for addressing the gaps and disruptions in 3D scene reconstruction. Nevertheless, as emphasized by \cite{gkitsas2021panodr, gao2022layout} traditional image inpainting techniques often struggle to preserve the structural integrity of indoor spaces. This can result in gaps that compromise the realism and completeness of the reconstructed scenes. In fact, when inpainting clutter-free indoor environments, the demands on preserving the surrounding context become even more stringent, aiming for true-to-life recreations rather than mere plausibility. To overcome this limitation, recent works such as PanoDR~\cite{gkitsas2021panodr} and LGPN~\cite{gao2022layout} have introduced the concept of conditioning the reconstruction process on layout maps. This conditioning step is essential for preserving structural information. However, both of these methods involve a two-phased approach, which includes layout estimation and image inpainting, and multiple subnetworks like a structural encoder, style encoder, and conditioned generator. Although these methods have proven their effectiveness, their efficiency for clutter-free room modeling is directly related to the robustness of the room layout estimation.
    
    In this work, we provide an innovate single-stage network capable of generating structurally coherent scenes without the need for structure conditioning. To achieve this, we draw on the unique features of indoor panoramas, which include symmetry and periodic structures, as well as the cyclic nature of panoramas. This understanding of indoor panorama characteristics serves as the foundation for our approach to improving the inpainting process for clutter-free room modeling. We take inspiration from recent work in image inpainting by Suvorov \etal~\cite{suvorov2022resolution} deemed LaMa. Their introduction of Fast Fourier Convolutions (FFCs) \cite{chi2020fast}, and more specifically, the Fourier transform within the image inpainting task has shown promise in handling human-made periodic structures. However the use of the Fourier transform in the context of indoor panorama settings for structure preservation has yet to be tested. Building on this inspiration, we propose a novel W-FourierMixer block specifically design for image inpainting in panorama settings. It differs from FFCs by applying the Fourier transform across height and width dimensions separately and leveraging gated convolutions for feature fusion.
    In addition to exploring the use of Fourier transform in a structure preserving panorama setting, we extend this exploration by integrating the proposed W-FourierMixer block into a more contemporary architectural framework, Uformer-like, drawing inspiration from Yu \etal~\cite{yu2022metaformer}, showing significant improvement LaMa. In summary, our contributions are as follows:
    \begin{itemize}
        \item We propose an innovative adversarial framework to transform a cluttered indoor environment into a clutter-free one using an Uformer-like generator and FourierFormer blocks. Several loss functions and new learning strategies using aggressive masks are proposed to improve model performance. Experiments demonstrate its superior performance compared to recent state-of-the-art models on Structured3d dataset~\cite{zheng2020structured3d}.
        \item We propose a novel W-FourierMixer block specifically design for image inpainting in panorama settings, leveraging Fourier transform and gated convolutions. Complementary ablation studies are conducted to assess the benefits of our W-FourierMixer block within our architecture.
    \end{itemize}

\section{Related Work}
\label{sec:related_work}
    \paragraph{}
    {\bf Clutter-free room modeling.}
    In clutter-free room modeling, the objective is to create precise and lifelike depictions of indoor spaces with minimal visual distractions. Various approaches have been proposed in the literature to generate clutter-free environments.
    Some methods involve the re-projection of actual background images acquired through prior \cite{queguiner2018towards, mori2015efficient} or simultaneous observations \cite{meerits2015real, kim2021generative} of the scene from different viewpoints. However, these approaches often require significant effort and specialized setups. Furthermore, these approaches fall short when certain portions of the surroundings are entirely blocked and unreachable. As a result, a more cost-efficient solution involves filling in or reconstructing backgrounds instead of attempting to recover the original ones.
    
    {\bf Image inpainting.}
    Traditional image inpainting approaches have historically relied on diffusion-based methods \cite{ballester2001filling}, which involve propagating pixels from neighboring regions to fill in the missing ones and patch-based methods \cite{hays2007scene,ding2018image,lee2016laplacian} which involves searching for and copying similar image patches from the rest of the image or existing image datasets.
    While traditional methods often yield visually realistic outcomes, their efficacy is contingent upon the simplicity of image structures. Without a comprehensive high-level understanding of the image contents, these methods encounter significant challenges, particularly when applied to images featuring intricate and complex structures.
    Initial data-driven inpainting methods combined autoencoders \cite{pathak2016context} with an adversarial loss \cite{creswell2018generative}, followed by other U-Net structure variants \cite{liu2018image, zeng2019learning}. To capture both local and global context, Iizuka \etal \cite{iizuka2017globally} employed dilated convolutions \cite{yu2015multi} with local and global discriminators. Additionally, \cite{liu2019coherent, xie2019image, yu2021diverse, li2022mat} utilized attention mechanisms, while \cite{suvorov2022resolution, lu2022glama, xu2023image, zheng2022image} applied frequency-enhanced transformations. To address the effects of masked inputs in feature extraction, Liu \etal \cite{liu2018image} proposed partial convolutions. Building on this, Yu \etal \cite{yu2019free} introduced a learnable gated mechanism called gated convolutions, and Yu \etal \cite{yu2020region} suggested region-wise normalization.
    Multi-stage generation \cite{yu2018generative} has garnered significant attention, as has the exploration of pluralistic generation \cite{jain2023keys, li2022mat, zhao2021large}, and large hole inpainting \cite{suvorov2022resolution, ma2022regionwise}.
    The impact of mask policies during training on model performance has also been explored. Surorov \etal \cite{suvorov2022resolution} adopted an aggressive mask policy, which was further extended by Lu \etal \cite{lu2022glama}, while Zheng \etal \cite{zheng2022image} implemented object-aware training.

    {\bf Boundary preserving image inpainting.}
    One crucial aspect of achieving photorealism in inpainting is the preservation of boundaries. Several approaches propose multi-stage methods that guide the inpainting process using intermediate cues such as salient edges \cite{nazeri2019edgeconnect} or semantic segmentation maps \cite{song2018spg}.
    In the context of indoor structural inpainting, Gkitsas \etal \cite{gkitsas2021panodr} conditions the generation by leveraging SEAN \cite{zhu2020sean} blocks and a 3-class semantic map corresponding to floor, ceiling, and wall labels. Additionally, Gkitsas \etal \cite{gkitsas2021panodr} employ a style encoder to extract texture information for each class. Building upon this, Gao \etal \cite{gao2022layout} extends the approach by incorporating HorizonNet\cite{sun2019horizonnet} to estimate layout edges, a 3-class segmentation, and plane segmentation, thereby enhancing boundary preservation and allowing for plane-wise style. It is important to note that both methods aim to improve boundary preservation in indoor modeling; however, they depend on the performance of the structure estimators.
    Our proposition differs in the sense that no intermediate cues are used to guide the generation.

    {\bf Fourier transform in neural networks.}
    The Fourier transform has been a cornerstone in image processing for many years. With the rise of deep learning in computer vision tasks, researchers have also explored their integration into neural networks.
    In \cite{lee2018single}, Fourier transform is applied in depth estimation, Yang \etal \cite{yang2020fda} focuses on segmentation, and Lee \etal \cite{lee2021fnet} employs it as a replacement for attention layers in NLP tasks while \cite{rao2021global} leverages it to learn long-term spatial dependencies in frequency domain, inspired by frequency filters in digital image processing. 
    The concept of Fast Fourier Convolutions (FFC) is introduced in \cite{chi2020fast}, consisting of both local and global branches. The global receptive field is achieved through Fourier Units, involving a point-wise convolution in the Fourier domain. Surorov \etal \cite{suvorov2022resolution} extend the use of FFC to image inpainting, demonstrating its excellence in recognizing repetitive patterns.
    While Fourier transform in FFCs is typically applied across height and width dimensions simultaneously, our work proposes a novel approach. We suggest applying two separate Fourier transforms for height and width, integrating them with gated convolutions and a windowed operation for improved performance.
\begin{figure*}[!h]
      \centering
        \includegraphics[width=0.95\linewidth]{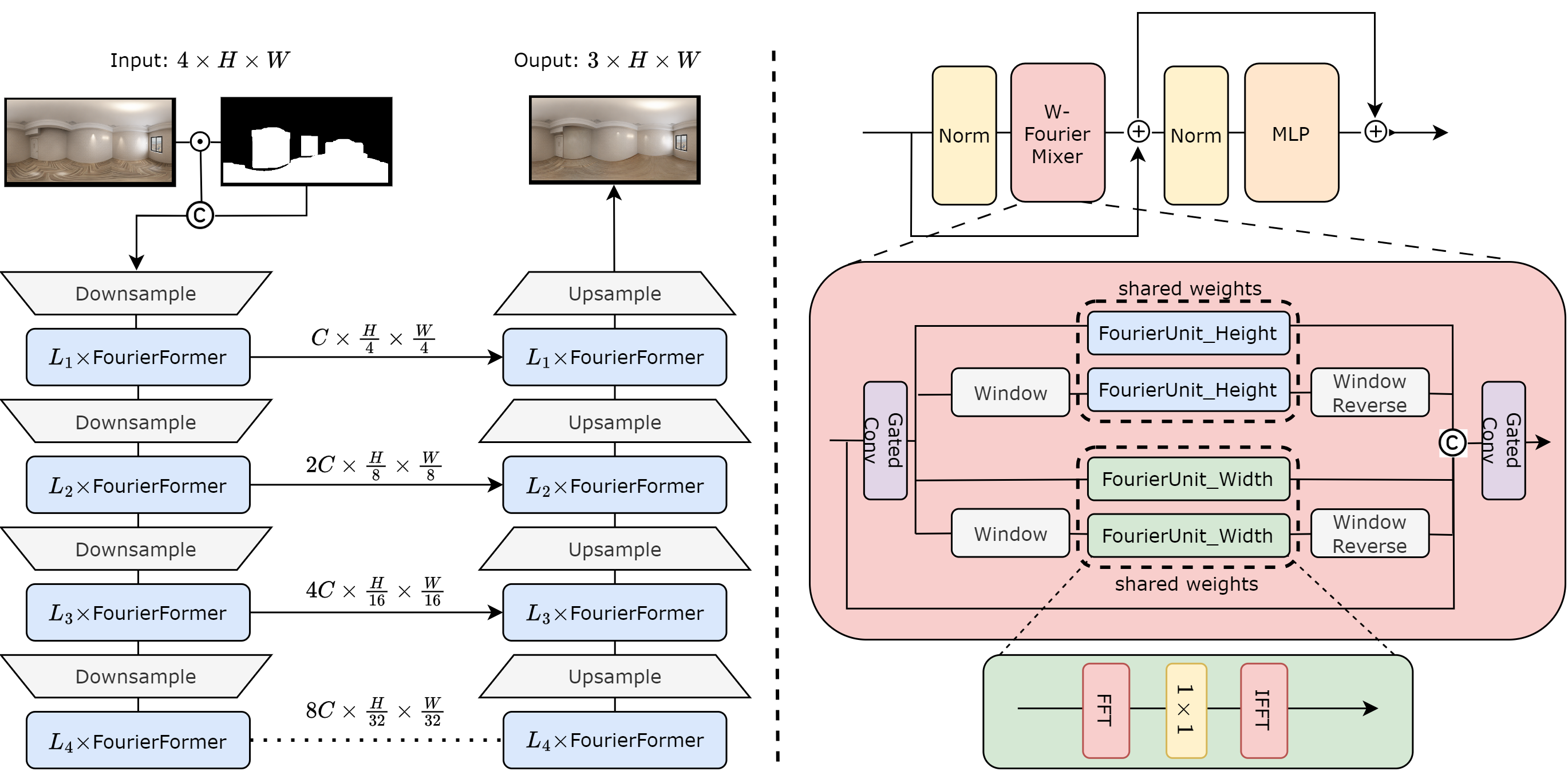}
    
       \caption{\textbf{Architecture overview.} Left: Given a furnished panoramic image of the interior scene as input and the mask of the objects to be removed, our approach generates a plausible empty scene. The proposed network architecture is based on an Uformer-like adversarial framework supervised by low- and high-level loss functions and a discriminator. Right: The architecture of the proposed W-FourierMixer block consists of Fourier Units applied separatly across height and width dimension enabling a large receptive field, and gated convolutions that facilitate a learnable gating mechanism. Window operations split the image in half before feeding it to the corresponding Fourier Unit allowing the capture of semi-local information. Fourier Units accross the same dimension share weights.
       }\label{fig:overview}
\end{figure*}

\section{Method}
\label{sec:method}

\paragraph{}
Our goal is to transform a cluttered indoor environment into a clutter-free one using an image inpainting approach. Our method takes as input $ x' = \mathcal{\text {\it concat}}(x \odot m, m) $, the concatenation of a masked indoor panorama and the corresponding binary mask and outputs a 3 channels color-image $\hat{x}$. We adopt an adversarial framework composed of a Uformer-like generator leveraging our proposed W-FourierMixer blocks $\mathcal {G}$ and a patch discriminator $\mathcal {D}$ \cite{isola2017image}. The training is supervised through a reconstruction loss on the unmasked pixels alongside a perceptual loss with a high receptive field \cite{suvorov2022resolution} and a non-saturating GAN loss \cite{goodfellow2014generative}. Our approach implements an aggressive mask policy, following \cite{suvorov2022resolution}, but with the proposal of additional masks to prove the effectiveness of the proposed method in several situations.

\subsection{Overall Architecture}
\label{subsubsec:Overall Architecture}

\paragraph{}
{\bf Generator.} The core of our model, illustrated in Figure~\ref{fig:overview}, adopts a U-shaped architecture comprising an encoder, a decoder, and skip connections. The encoder features four hierarchical stages, each equipped with a downsample layer and \(L_i\) FourierFormer blocks Yu \etal Metaformer blocks \cite{yu2022metaformer}. These blocks introduce a feature dimension \(D_i \in \mathbb{R}^{{2^{i-1}C} \times \frac{H}{2^{(i+1)}} \times \frac{W}{2^{(i+1)}}}\) for \(i \in [1,4]\). Downsampling is achieved through convolutional layers, with the first layer having a kernel size of 7 and a stride of 4 (scaling factor of \(\frac{1}{4}\)), followed by subsequent layers with a kernel size of 3 and a stride of 2 (scaling factor of \(\frac{1}{2}\)). The symmetrically aligned decoder mirrors the encoder, featuring \(L_i\) FourierFormer blocks and an upsample layer for each stage.
In accordance with established practices in \cite{yu2022metaformer}, our implementation incorporates Layer Normalization and StarReLU activations, with the exclusion of biases.
Given the inherently spherical nature of our input data, we apply circular padding \cite{sun2019horizonnet} in the horizontal image direction and reflection padding in the vertical direction. This strategic padding helps overcome longitudinal boundary discontinuities and simulates singularities at the poles \cite{zioulis2021single}.

{\bf FourierFormer.} Motivated by advancements in transformer-like convolutional architectures \cite{yu2022metaformer}, our network incorporates FourierFormer blocks, depicted Figure~\ref{fig:overview} right top corner, as fundamental components. These blocks amalgamate spatial mixing and channel mixing modules, enhancing feature representation. The block equations are defined as:
\begin{equation}
    X' = X + \text{{token\_mixer}}(\text{{norm}}(X))
\end{equation}
\begin{equation}
    X' = X' + \text{{channel\_mixer}}(\text{{norm}}(X')),
\end{equation}
where \(X\) denotes the input feature map, \(\text{{norm}}(\cdot)\) signifies a normalization layer, \(\text{{token\_mixer}}(\cdot)\) denotes the spatial mixing operation, and \(\text{{channel\_mixer}}(\cdot)\) represents the channel mixing operation. This configuration enables long-range dependency aggregation and inter-channel relationships.

In our method, we leverage our proposed W-FourierMixer as the \text{{token\_mixer}} and a multi-layer perceptron (MLP) composed of two pointwise convolutions with an activation in between as the \text{{channel\_mixer}}.

{\bf W-FourierMixer.}
The proposed W-FourierMixer, depicted in Figure~\ref{fig:overview}, consists of two crucial components: two Fourier Units applying the Fast Fourier Transform (FFT) independently across the height or width dimension, and gated convolutions.
The first gated convolution is employed for reducing the input feature dimensions by half, while the Fourier Units contribute to global receptive field expansion and the detection of repetitive patterns \cite{jain2023keys}, and finally the gated convolutions acts as a masked informed feature fusion.

A notable departure from \cite{chi2020fast} is W-FourierMixer's distinct approach to FFT, performing it separately across the height and width dimensions. Experimental results, as presented in the supplemental material, reveal that the application of convolution in the Fourier domain with Fourier Units leads to the emergence of a symmetry effect. In contrast to the original Fast Fourier Convolutions (FFC) that result in cross symmetry, W-FourierMixer achieves either vertical or horizontal symmetry by individually applying FFT across height or width. This proves advantageous for tasks involving indoor scenes where vertical and horizontal symmetry is prevalent, allowing the model to effectively capture features such as horizon/ceiling height lines and cuboid-shaped rooms.
An additional refinement involves a window operation, where the input is split into two parts before feeding it to the Fourier Unit layer, facilitating symmetry in a more localized manner. Crucially, the Fourier Unit used for the windowed feature map is shared with the one used for the entire feature map, enabling weight-sharing within the same layer.

To facilitate feature fusion within W-FourierMixer, we employ Gated Convolutions \cite{yu2019free}, a technique that extends the concept of partial convolutions \cite{liu2018image}. Gated Convolutions introduce a learnable gating mechanism for selecting relevant features, enhancing the model's ability to gradually attend to unasked regions in a learnable fashion.

\subsection{Loss Function}
\label{subsubsec:loss}
    \paragraph{}
    Leveraging a synthetic dataset composed of cluttered/clutter-free image pairs, we employ supervised training.
    The model is supervised under a combination of losses:
    \begin{equation}
        \begin{split}
            \mathcal {L}_{\text {\it Final}} = & 
            \lambda_{\text {\it Rec}} \mathcal {L}_{\text {\it Rec}}
            + 
            \lambda_{\text {\it Perc}} \mathcal {L}_{\text {\it Perc}} 
            + 
            \lambda_{\text {\it Adv}} \mathcal {L}_{\text {\it Adv}}\\
            &+ 
            \lambda_{\text {\it GP}} \mathcal {L}_{\text {\it GP}}
            + 
            \lambda_{\text {\it FM}} \mathcal {L}_{\text {\it FM}}
        \end{split}
    \end{equation}

    { \bf Reconstruction loss.}
    Unmasked pixels should not be changed during the inpainting process. To enforce such constraint, we apply a reconstruction loss ($L_1$) between unmasked pixels:
    \begin{equation}
        \mathcal {L}_{\text {\it Rec}}(x, \hat x) = \| x - \hat x \|_1 \odot (1-m),
    \end{equation}
    where \(x\) represents ground truth and \(\hat x\) represents the inpainting network prediction.

    { \bf Perceptual loss.}
    Applying a reconstruction loss on the masked pixels constraints the generator to reconstruct the ground truth precisely. This approach exhibits limitations, particularly when dealing with large masks and insufficient  context from the unmasked image. In such cases, the reconstruction loss tends to produce blurry results, as the generator learns an average representation of plausible outcomes.
    To overcome these limitations, we adopt a perceptual loss encouraging the generated image and the ground truth to have similar characteristics while allowing for variations in the reconstruction.
    Perceptual loss evaluates the distance between features extracted from the predicted and target images using a pre-trained base network, $\phi(\cdot)$. Since our task focuses on understanding the global structure --~the room layout~-- we employ a High Receptive Field Perceptual Loss:
    \begin{equation}
       \mathcal {L}_{\text {\it Perc}}(x, \hat {x}) = \mathcal {M}([\phi _{\text {\it HRF}}(x)-\phi _{\text {\it HRF}}(\hat {x})]^2)
    \end{equation}
    where $[\cdot - \cdot]^2$ is an element-wise operation, $\mathcal {M}$ is the sequential two-stage mean operation (interlayer mean of intra-layer means) and $\phi(\cdot)$ is a dilated ResNet-50 \cite{he2016deep} pretrained for ADE20K \cite{zhou2017scene} semantic segmentation.

    { \bf Adversarial loss.}
    To make sure our inpainting network creates plausible indoor images, we adopt a non-saturating adversarial loss, denoted as $\mathcal {L}_{\text {\it Adv}}$. 
    This loss is integrated using a patch-based discriminator, which aims to refine the inpainting process by learning an adaptive loss. The discriminator is trained to distinguish between real and inpainted images.
    Unlike traditional discriminators that provide a single output indicating a true or false prediction, a patch-based discriminator gives a tensor of logits, representing true or false patches. This means it evaluates realism not globally, but in localized areas of the image.
    Adversarial loss is defined as:
    \begin{equation}
        \mathcal {L}_{\text {\it Adv}} =  \mathcal {L}_{\text {\it D}} + \mathcal {L}_{\text {\it G}},
    \end{equation}
    where
    \begin{equation}
        \begin{split}
        \mathcal {L}_{\text {\it D}} = 
        & -\mathbb{E}_x[\log \mathcal {D}(x)] \\
        & - \mathbb{E}_{\hat{x},m}[\log \mathcal {D}(\hat{x}) \odot (1 - m)] \\
        & - \mathbb{E}_{\hat{x},m}[\log(1 - \mathcal {D}(\hat{x})) \odot m]
        \end{split}
    \end{equation}
    represents the discriminator loss and
    \begin{equation}
        \mathcal {L}_{\text {\it G}} = -\mathbb{E}_{\hat{x}}[\log D(\hat{x})]
    \end{equation}
    represents the generator loss, where only masked patches are regarded as fake.

    Additionally, we use a feature matching loss $\mathcal {L}_{\text {\it FM}}$ \cite{DBLP:journals/corr/abs-1711-11585} which ensures the generated images match the statistics of the ground truth by evaluating the distance between discriminator's features for true and false samples (it is in fact a discriminator perceptual loss); and we use gradient penalty $\mathcal {L}_{\text {\it GP}}$ ensuring the discriminator cannot
    create a non-zero gradient orthogonal to the data distribution without suffering a loss.
    \begin{equation}
        \mathcal {L}_{\text {\it GP}} = \mathbb{E}_x \|\nabla_x D(x)\|^2
    \end{equation}
    \begin{equation}
        L_{\text{FM}} = \| \psi(\mathcal {D}(x)) - \psi(\mathcal {D}(\hat x)) \|^2
    \end{equation}
    where $\psi$ is the feature extraction operation.

\subsection{Mask policy}
\label{subsubsec:masks}
    \paragraph{}
    We apply masks to empty scenes and use cluttered scenes only for semantic annotations. This decision is driven by the distinct lighting settings in ray-traced renderings of cluttered and clutter-free scenes, which may cause irregular supervision across data samples.
    
    In the same way that data augmentation is crucial for improving discriminative models, the effectiveness of the inpainting network heavily depends on the approach to mask generation \cite{sun2019horizonnet, lu2022glama, zheng2022image}. To enhance training, we adopt an aggressive mask generation strategy based on \cite{suvorov2022resolution}.

\begin{figure*}[h!]
    \centering  
    \includegraphics[width=1.\linewidth]{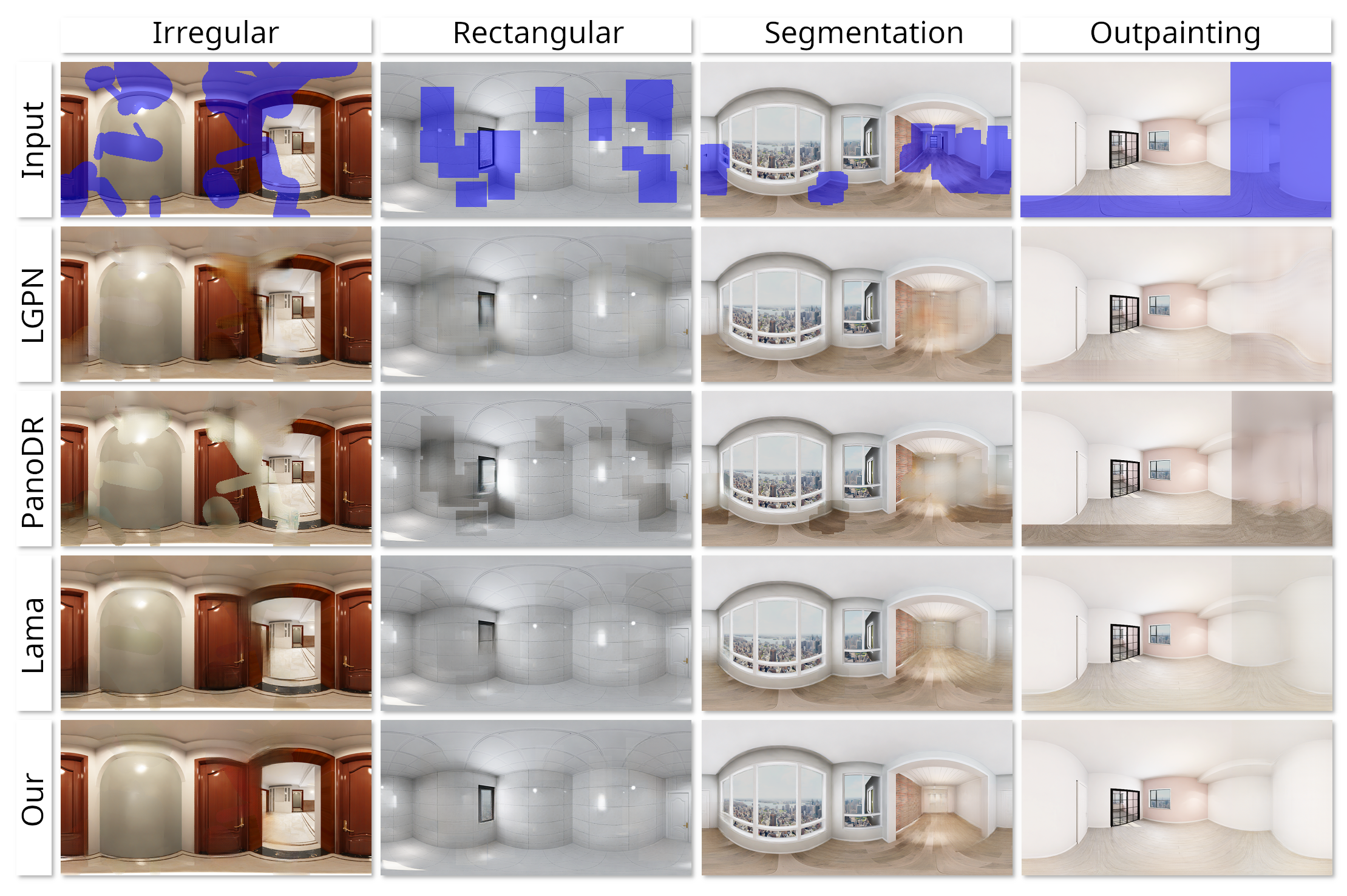}  
    \caption{\textbf{Qualitative results with state-of-the-arts for transforming a cluttered indoor environment into a clutter-free room.} The columns represent the different masks with more or less important ratios. The rows represent the approaches of the literature. The results show the effectiveness of our approach regardless of mask and room properties.}
    \label{fig:qualitative_comparison}
\end{figure*}

    {\bf Mask Types.}
    Expanding on Surorov \etal \cite{suvorov2022resolution} approach, we randomly sample masks during training from fives types. These include polygonal strips dilated by random strokes ('Irregular') and rectangles ('Rectangular') with arbitrary aspect ratios, following \cite{suvorov2022resolution}. Additionally, we introduce "Segmentation," "Outpainting," "Quadrants," as illustrated in Figure \ref{fig:qualitative_comparison}.

    {\bf Segmentation Masks.}
    Utilizing a synthetic indoor dataset with pixel-perfect semantic annotations, we create semantic masks representing clutter. Applying dilation kernels to the mask emulates user input, simulating noisy semantic segmentation predictions from other neural networks. Adapting the number of dilation iterations based on clutter volume ensures a more aggressive mask with a higher ratio of masked pixels. This approach proves superior at producing semantic shaped masks compared to Gkitsas \etal \cite{gkitsas2021panodr} approach as shown in supplementary material, which focuses on masking convex shapes.
    
    {\bf Outpainting and Quadrants Masks.}
    In addition to the aforementioned masks, we generate outpainting masks, where an entire outward portion of the image is masked, and quadrant masks, where only an outward subportion is masked. These mask types are instrumental in training the model to generate plausible structures across image borders, further enhancing its ability to handle diverse inpainting scenarios.

\section{Experiments}
\label{sec:results}


\subsection{Dataset and Metrics}
\label{subsubsec:dataset_metrics}
\paragraph{}
{\bf Dataset.}
Experiments were conducted on the public indoor panorama dataset, Structured3D ~\cite{zheng2020structured3d}, consisting of 21,835 indoor panoramas with three different configurations (empty, simple, and full room) as well as layout and semantic label annotations, making it highly suitable for structural inpainting. We adopt the official data split, resulting in 18,362 scenes for training, 1,776 for validation, and 1,697 for testing.

{\bf Metrics.}
Given the generative nature of our task, which aims at indoor structural inpainting, the generation of plausible and photo-realistic content is considered a crucial objective. Our focus is to preserve layout structure: consequently we employ full-reference evaluation metrics, including the Mean Absolute Error (MAE), the Peak Signal-to-Noise Ratio (PSNR), and the Structural Similarity Index (SSIM), to evaluate performance.

\subsection{Implementation Details}
\label{subsubsec:implementation}
\paragraph{}
{\bf Network Details.}
The parameter \(C\) is configured to be 64, and \(L_1, L_2, L_3, L_4\) are set to 2.

{\bf Training settings.}
We implemented our model in PyTorch and conducted experiments on a single GeForce RTX 3060 GPU with 12GB VRAM.  We employed the AdamW optimizer \cite{loshchilov2018decoupled} with fixed learning rates of 0.001 and 0.0001 for the inpainting and discriminator networks, respectively. We empirically set $\lambda_{\text {\it Rec}}=10, \lambda_{\text {\it Perc}}=100, \lambda_{\text {\it Adv}}=10, \lambda_{\text {\it GP}}=0.001$ and $\lambda_{\text {\it FM}}=30$.
The final model, with dimensions of \(512 \times 256\), is trained for 40 epochs using a batch size of 6. 

\begin{table*}[!t]
\centering
\resizebox{\textwidth}{!}{%
\begin{tabular}{llllll|llll|llll|}
\toprule
\multicolumn{2}{c}{\multirow{2}{*}{\large \bf Mask}} & \multicolumn{4}{c}{\large \bf MAE $\downarrow$} & \multicolumn{4}{c}{\large \bf PSNR $\uparrow$} & \multicolumn{4}{c}{\large \bf SSIM $\uparrow$}
\\ \cmidrule(lr){3-6} \cmidrule(lr){7-10} \cmidrule(lr){11-14} 
       &   & Ours & LaMa & LGPN & PanoDR & Ours & LaMa & LGPN & PanoDR & Ours & LaMa & LGPN & PanoDR \\
\cmidrule(lr){1-2} \cmidrule(lr){3-6} \cmidrule(lr){7-10} \cmidrule(lr){11-14} 
\multicolumn{1}{l|}{\multirow{5}{*}{\rotatebox[origin=c]{90}{Segmentation}}} & \multicolumn{1}{l|}{\hspace{2mm}1-10\%} & \textbf{0.0029} & 0.0035 & 0.0031 & 0.0151 & \textbf{37.673} & 36.278 & 36.482 & 28.287 & \textbf{0.9826} & 0.9801 & 0.9810 & 0.9363 \\
\multicolumn{1}{l|}{} & \multicolumn{1}{l|}{10-20\%} & \textbf{0.0069} & 0.0082 & 0.0075 & 0.0236 & \textbf{32.412} & 31.261 & 31.480 & 25.237 & \textbf{0.9616} & 0.9568 & 0.9588 & 0.9073 \\
\multicolumn{1}{l|}{} & \multicolumn{1}{l|}{20-30\%} & \textbf{0.0093} & 0.0114 & 0.0113 & 0.0321 & \textbf{31.486} & 30.122 & 29.509 & 23.568 & \textbf{0.9466} & 0.9383 & 0.9359 & 0.8694 \\
\multicolumn{1}{l|}{} & \multicolumn{1}{l|}{30-40\%} & \textbf{0.0187} & 0.0222 & 0.0217 & 0.0445 & \textbf{26.715} & 25.737 & 25.485 & 21.407 & \textbf{0.9091} & 0.8998 & 0.9001 & 0.8402 \\
\multicolumn{1}{l|}{} & \multicolumn{1}{l|}{40-50\%} & \textbf{0.0259} & 0.0303 & 0.0303 & 0.0554 & \textbf{24.951} & 24.069 & 23.687 & 20.149 & \textbf{0.8824} & 0.8722 & 0.8697 & 0.8081 \\ 
\cmidrule(lr){1-2} \cmidrule(lr){3-6} \cmidrule(lr){7-10} \cmidrule(lr){11-14} 
\multicolumn{1}{l|}{\multirow{5}{*}{\rotatebox[origin=c]{90}{Irregular}}} & \multicolumn{1}{l|}{1-10\%} & \textbf{0.0024} & 0.0031 & 0.0030 & 0.0166 & \textbf{38.801} & 37.069 & 36.324 & 27.612 & \textbf{0.9854} & 0.9828 & 0.9821 & 0.9344 \\
\multicolumn{1}{l|}{} & \multicolumn{1}{l|}{10-20\%} & \textbf{0.0052} & 0.0063 & 0.0065 & 0.0252 & \textbf{33.554} & 33.113 & 32.180 & 24.981 & \textbf{0.9700} & 0.9649 & 0.9629 & 0.9066 \\
\multicolumn{1}{l|}{} & \multicolumn{1}{l|}{20-30\%} & \textbf{0.0082} & 0.0102 & 0.0113 & 0.0355 & \textbf{31.710} & 30.416 & 29.094 & 22.969 & \textbf{0.9520} & 0.9441 & 0.9384 & 0.8729 \\
\multicolumn{1}{l|}{} & \multicolumn{1}{l|}{30-40\%} & \textbf{0.0118} & 0.0146 & 0.0170 & 0.0473 & \textbf{29.692} & 28.504 & 26.877 & 21.389 & \textbf{0.9319} & 0.9209 & 0.9100 & 0.8368 \\
\multicolumn{1}{l|}{} & \multicolumn{1}{l|}{40-50\%} & \textbf{0.0163} & 0.0201 & 0.0247 & 0.0607 & \textbf{27.884} & 26.766 & 24.852 & 20.025 & \textbf{0.9084} & 0.8942 & 0.8750 & 0.7975 \\
\cmidrule(lr){1-2} \cmidrule(lr){3-6} \cmidrule(lr){7-10} \cmidrule(lr){11-14} 
\multicolumn{1}{l|}{\multirow{5}{*}{\rotatebox[origin=c]{90}{Rectangular}}} & \multicolumn{1}{l|}{1-10\%} & \textbf{0.0028} & 0.0036 & 0.0032 & 0.0168 & \textbf{37.484} & 35.709 & 35.835 & 27.301 & \textbf{0.9832} & 0.9801 & 0.9797 & 0.9313 \\
\multicolumn{1}{l|}{} & \multicolumn{1}{l|}{10-20\%} & \textbf{0.0052} & 0.0067 & 0.0062 & 0.0239 & \textbf{33.844} & 32.340 & 32.202 & 25.068 & \textbf{0.9693} & 0.9637 & 0.9625 & 0.9054 \\
\multicolumn{1}{l|}{} & \multicolumn{1}{l|}{20-30\%} & \textbf{0.0091} & 0.0115 & 0.0115 & 0.0351 & \textbf{30.694} & 29.402 & 28.772 & 22.773 & \textbf{0.9477} & 0.9384 & 0.9341 & 0.8656 \\
\multicolumn{1}{l|}{} & \multicolumn{1}{l|}{30-40\%} & \textbf{0.0137} & 0.0170 & 0.0182 & 0.0472 & \textbf{28.455} & 27.287 & 26.300 & 21.113 & \textbf{0.9229} & 0.9104 & 0.9005 & 0.8231 \\
\multicolumn{1}{l|}{} & \multicolumn{1}{l|}{40-50\%} & \textbf{0.0179} & 0.0221 & 0.0248 & 0.0587 & \textbf{26.455} & 25.884 & 24.616 & 19.897 & \textbf{0.9012} & 0.8865 & 0.8704 & 0.7869 \\
\cmidrule(lr){1-2} \cmidrule(lr){3-6} \cmidrule(lr){7-10} \cmidrule(lr){11-14} 
\multicolumn{1}{l|}{\multirow{5}{*}{\rotatebox[origin=c]{90}{Outpainting}}} & \multicolumn{1}{l|}{1-10\%} & \textbf{0.0025} & 0.0032 & 0.0033 & 0.0152 & \textbf{40.009} & 38.094 & 37.259 & 28.769 & \textbf{0.9880} & 0.9863 & 0.9862 & 0.9478 \\
\multicolumn{1}{l|}{} & \multicolumn{1}{l|}{10-20\%} & \textbf{0.0065} & 0.0079 & 0.0095 & 0.0245 & \textbf{34.096} & 32.834 & 30.753 & 25.663 & \textbf{0.9698} & 0.9669 & 0.9661 & 0.9251 \\
\multicolumn{1}{l|}{} & \multicolumn{1}{l|}{20-30\%} & \textbf{0.0128} & 0.0151 & 0.0185 & 0.0367 & \textbf{29.695} & 28.589 & 26.656 & 23.005 & \textbf{0.9456} & 0.9412 & 0.9386 & 0.8956 \\
\multicolumn{1}{l|}{} & \multicolumn{1}{l|}{30-40\%} & \textbf{0.0207} & 0.0240 & 0.0293 & 0.0508 & \textbf{26.634}  & 25.800 & 23.796 & 21.048 & \textbf{0.9211} & 0.9160 & 0.9088 & 0.8656 \\
\multicolumn{1}{l|}{} & \multicolumn{1}{l|}{40-50\%} & \textbf{0.0284} & 0.0317 & 0.0398 & 0.0615 & \textbf{24.710} & 24.184 & 22.145 & 20.066 & \textbf{0.8944} & 0.8884 & 0.8766 & 0.8373 \\
\cmidrule(lr){1-2} \cmidrule(lr){3-6} \cmidrule(lr){7-10} \cmidrule(lr){11-14} 
\multicolumn{1}{l|}{\multirow{5}{*}{\rotatebox[origin=c]{90}{Quadrants}}} & \multicolumn{1}{l|}{1-10\%} & \textbf{0.0020} & 0.0027 & 0.0030 & 0.0144 & \textbf{41.829} & 39.379  & 38.571 & 29.259 & \textbf{0.9888} & 0.9875 & 0.9878 & 0.9509 \\
\multicolumn{1}{l|}{} & \multicolumn{1}{l|}{10-20\%} & \textbf{0.0064} & 0.0077 & 0.0090 & 0.0244 & \textbf{34.082} & 32.680 & 31.065 & 25.372 & \textbf{0.9697} & 0.9665 & 0.9668 & 0.9252 \\
\multicolumn{1}{l|}{} & \multicolumn{1}{l|}{20-30\%} & \textbf{0.0133} & 0.0158 & 0.0184 & 0.0377 & \textbf{29.066} & 27.944 & 26.495 & 22.429 & \textbf{0.9421} & 0.9373 & 0.9370 & 0.8903 \\
\multicolumn{1}{l|}{} & \multicolumn{1}{l|}{30-40\%} & \textbf{0.0211} & 0.0252 & 0.0284 & 0.0516 & \textbf{26.138} & 25.126 & 23.908 & 20.501 & \textbf{0.9139} & 0.9076 & 0.9062 & 0.8559 \\
\multicolumn{1}{l|}{} & \multicolumn{1}{l|}{40-50\%} & \textbf{0.0302} & 0.0356 & 0.0403 & 0.0664 & \textbf{23.870} & 23.041 & 21.915 & 18.999 & \textbf{0.8833} & 0.8759 & 0.8702 & 0.8184 \\ \bottomrule
\end{tabular}%
}
\caption{Quantitative evaluation on Structured3d dataset\cite{zheng2020structured3d}, for different types of masks, e.g., Segmentation, Irregular, Rectangular, Outpainting and Quadrants, and different mask ratio from 10\% to 50\%. Approaches are evaluated according to three metrics. $\downarrow$ indicates that the lower the result, the better it is and vice versa. The best results by mask and metric are shown in bold.}
\label{tab:quantitative_results}
\end{table*}

{\bf Baselines.}
We compare our model against state-of-the-art methods on the Structured3D dataset ~\cite{zheng2020structured3d}, specifically PanoDR~\cite{gkitsas2021panodr} and LGPN~\cite{gao2022layout}. Additionally, we compare against LaMa~\cite{suvorov2022resolution}, as it also utilizes the Fourier Domain for image inpainting. Pretrained models on the same dataset for PanoDR and LGPN are used, and LaMa is trained under the same settings as our method at $512\times 256$ resolution.

{\bf Evaluation settings.}
We evaluate our method across fives mask types and at five different mask ratio intervals, stopping at 50\%. As detailed in Section~\cref{subsubsec:masks}, segmentation masks are dilated until the minimum interval ratio is reached. If no segmentation candidate is available (empty room), random rectangular masks are created.

\subsection{Results and Comparisons}
\label{subsubsec:results_comparisons}
\paragraph{}
{\bf Qualitative Results.} We compare our approach with the following state-of-the-art approaches PanoDR~\cite{gkitsas2021panodr}, LGPN~\cite{gao2022layout} and LaMa~\cite{suvorov2022resolution}. To assess the performance of these approaches, Figure \ref{fig:qualitative_comparison} illustrates several examples, with different masks and rooms. 
In terms of performance, inpainting-based approaches e.g., LaMa and ours, provide more realistic results, both in terms of texture and structure. Approaches based on layout estimates are more difficult, especially in the presence of complex perspectives, with depths of field varying from one side of the room to the other, as in columns 1 and 3. Compared with LaMa, the proposed approach better preserves the environmental shade, as well as repetitive patterns such as floor and wall tiles, and room perspectives. This major difference is brought about by the proposed W-FourierMixer block which consists of Fourier Units applied separately across height and width dimension enabling a large receptive field, and gated convolutions that facilitate a learnable gating mechanism.

{\bf Quantitative Results.}
Table \ref{tab:quantitative_results} shows models scores for full-reference objectives. The results highlight the performance of the proposed approach in comparison with recent approaches in the literature. Both on the different masks, with more or less important ratios, and on the different metrics, our approach outperforms other approaches. In terms of performance, our approach performs better than LaMa, demonstrating the contribution of our proposed W-FourierMixer block compared with a Fast Fourier convolution. The various metrics highlight several quality factors of our approach, including the ability to faithfully reconstruct pixels in terms of contrast, brightness and color.

\subsection{Ablation Study}
\label{subsubsec:Ablation}

Here, we conduct ablation studies to validate our model from different angles. We first evaluate the contribution of different token mixers within a U-Former like architecture, followed by a study on the window operation and finally the impact of the chosen perceptual loss on the model performance.

{\bf Token Mixers.}
\Cref{tab:token_mixer_ablation} shows the contribution of the proposed W-FourierMixer against: \(i\)) FFCs, \(ii\)) Gated Convolutions, \(iii\)) 2D W-FourierMixer(2D-WFM) where the Fourier transform is applied accross the 2 dimensions simultaneously as in FFCs. Results demonstrate that the combination of gated convolutions with the 2d Fourier transform alone, outperforms FFCs and Gated Convolutions. As seen by the difference in MAE metric between UFormer 2D-FMW and our method, applying Fourier transform separately improves performance resulting in finer contours.

{\bf Window Operation.}
As demonstrated by \Cref{tab:window_ablation}, the window operation is indeed crucial for the model performance. We deemed it to be beneficial at rooms like the Irregular or Segmentation examples in ~\cref{fig:qualitative_comparison} where the more symmetric information is localized in half of the image.

{\bf Perceptual loss.}
\Cref{tab:loss_ablation} reports metrics for the proposed model trained with a Low Receptive Field Perceptual Loss (LRFPL), such as a pretrained VGG19, and a High Receptive Field Perceptual (HRFP), such as a pretrained Dilated ResNet50. We find that the high receptive field loss is a key component for comprehending the global structure of the room.

\begin{table}[]
\centering
\resizebox{\columnwidth}{!}{%
\begin{tabular}{llll}
\toprule
Architectures & MAE $\downarrow$ & PSNR $\uparrow$ & SSIM $\uparrow$ \\
\midrule
Uformer FFC        & 0.0098 & 30.953 & 0.9356 \\
Uformer Gated Conv & 0.0103 & 30.505 & 0.9335 \\
Uformer 2D-WFM     & 0.0095 & 31.494 & 0.9401 \\
Uformer WFM(ours)  & \textbf{0.0089} & \textbf{31.785} & \textbf{0.9414} \\
\bottomrule
\end{tabular}%
}
\caption{Ablation study over token mixer blocks. $\downarrow$ indicates that the lower the result, the better it is and vice versa. The best results are shown in bold.}
\label{tab:token_mixer_ablation}
\end{table}

\begin{table}[]
\centering
\resizebox{\columnwidth}{!}{%
\begin{tabular}{llll}
\toprule
Architectures & MAE $\downarrow$ & PSNR $\uparrow$ & SSIM $\uparrow$ \\
\midrule
Uformer FM & 0.0098 & 30.861 & 0.9360 \\
Uformer WFM(ours) & \textbf{0.0089} & \textbf{31.785} & \textbf{0.9414} \\
\bottomrule
\end{tabular}%
}
\caption{Ablation study over window operation. $\downarrow$ indicates that the lower the result, the better it is and vice versa. The best results are shown in bold.}
\label{tab:window_ablation}
\end{table}

\begin{table}[]
\centering
\resizebox{\columnwidth}{!}{%
\begin{tabular}{llll}
\toprule
Architectures & MAE $\downarrow$ & PSNR $\uparrow$ & SSIM $\uparrow$ \\
\midrule
Uformer WFM LRFPL & 0.0098 & 30.969 & 0.9354 \\
Uformer WFM HRFPL (ours) & \textbf{0.0089} & \textbf{31.785} & \textbf{0.9414} \\
\bottomrule
\end{tabular}%
}
\caption{Ablation study over the perceptual loss. $\downarrow$ indicates that the lower the result, the better it is and vice versa. The best results are shown in bold.}
\label{tab:loss_ablation}
\end{table}
\section{Conclusion}
\label{sec:conclusion}

Modeling a clutter-free room from a single image without relying on constraining layout estimators to guide the reconstruction process is a challenging task. To deal with this issue, we have proposed an innovative approach inspired by recent architectures such as Metaformers and frequency enhanced operations such as FFCs. To optimize the performance of existing models, we have created a new token mixer block based on Fourier transfom called Windowed-FourierMixer (W-FM). In contrast to the original Fast Fourier Convolutions (FFC) that result in cross symmetry, W-FM achieves either vertical or horizontal symmetry by individually applying FFT across height or width. Further refinement through a windowed feature map, and a gated convolution is employed to improve reconstruction performance. The evaluations highlight the performance of the proposed approach in comparison with recent approaches in the literature. The ablation study consolidates our study by assessing the contribution of W-FM to the reconstruction process.

\paragraph{Limitations.} In this work, we tested our approach only on synthetic indoor scenes, but real-world scenes have a complex lighting and layout structure that makes it more difficult to remove elements. However, the amount of real-world scene data and the quality of furniture category annotations are insufficient to train our model, so we chose to train it only on the Structured3D synthetic dataset.

\paragraph{Future work.} While our work aims at clutter-free room modeling, our contributions stop at the end of the inpainting process. Further improvements could be brought to the network such as a complementary depth prediction, or other supervision losses such as a perceptual loss on a Room Layout Estimation Network or a 3d informed depth loss.
{
    \small
    \bibliographystyle{ieeenat_fullname}
    \bibliography{main}
}

\clearpage
\setcounter{page}{1}
\maketitlesupplementary

\section{Fourier Unit Features}
\label{sec:fourierunit_features}

In this section, we explore the effects of the Fourier Unit corresponding to a pointwise convolution in the Fourier domain. 

\begin{figure}
    \centering  
    \includegraphics[width=1.0\columnwidth]{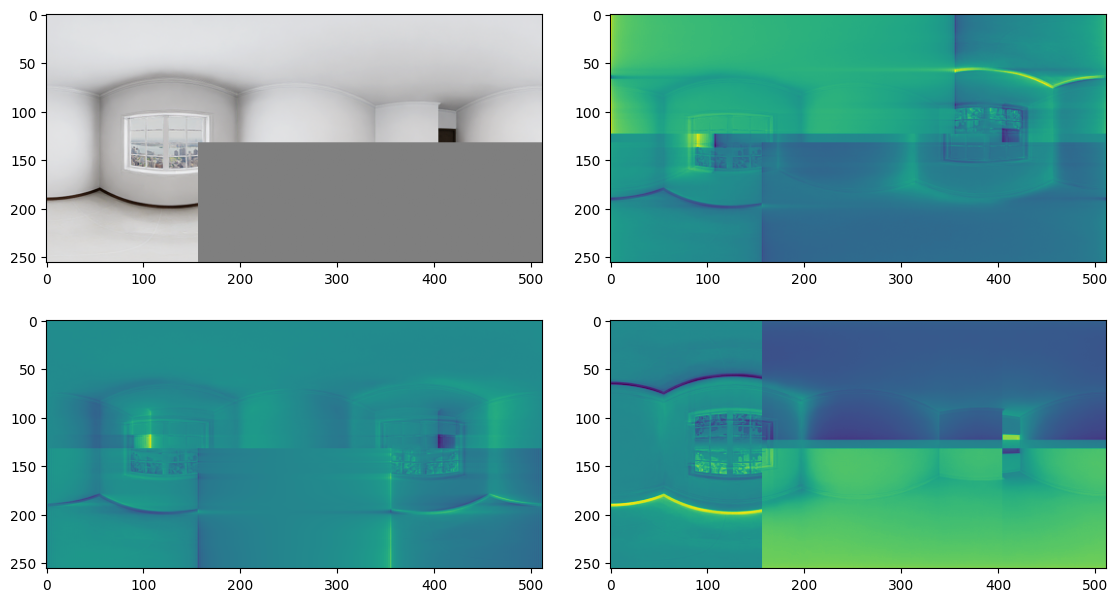}  
    \caption{\textbf{Fourier unit outputs across different dimensions.} The top-left image represents the input, the top-right shows the output on the Fourier transform across height and width, the bottom-left on width alone, and the bottom-right on height alone. Only the first channel of the output is displayed for better visualization; no activation or normalization was applied after the convolution.}
    \label{fig:fourierFeats1}
\end{figure}

Figure \ref{fig:fourierFeats1} visually demonstrates the impact of applying a pointwise convolution in the Fourier transform of both width and height dimensions, resulting in cross-symmetry. Conversely, applying it solely on the width yields symmetry across a vertical axis, while applying it only on the height results in symmetry across a horizontal axis.

\begin{figure}
    \centering  
    \includegraphics[width=1.0\columnwidth]{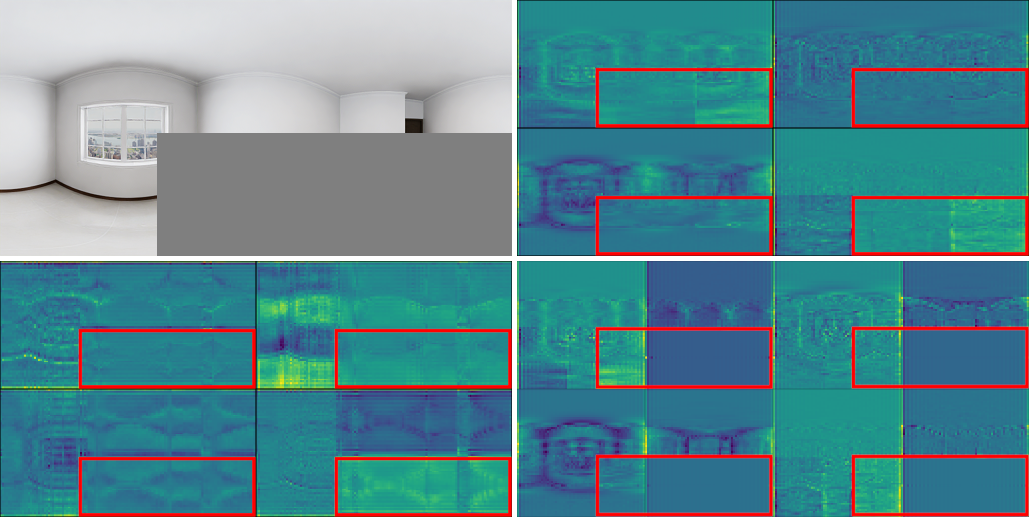}  
    \caption{\textbf{Features from Fourier units in the first block of the first stage of the encoder.} The top-left image represents the network input, the top-right shows the output from the Fourier unit across width, the bottom-left across height, and the bottom-right the corresponding features across width with a window operation. Indeed, the symmetry effect is present in the feature maps, allowing the network to grasp a global structure understanding right from the early stages.}
    \label{fig:qFourierFeats2}
\end{figure}

We extend this exploration to our network, focusing on the first block of the first stage of the encoder. In Figure \ref{fig:qFourierFeats2}, output features from the Fourier units reveal a symmetrical population of masked regions with information from the non-masked regions. This approach enables the network to capture global structural understanding from the initial stages. Additionally, the use of window operations contributes to local-scale symmetry.

\section{Masks and Ratios}
\label{sec:masks_and_ratios}
In this section, we delve into the statistics of clutter within the Structured3D dataset. Our approach involves calculating the amount of clutter per scene, defined as the ratio or percentage of pixels associated with furniture and other non-structural labels in the semantic maps for each image.
\begin{figure}[b]
    \centering  
    \includegraphics[width=1.0\linewidth]{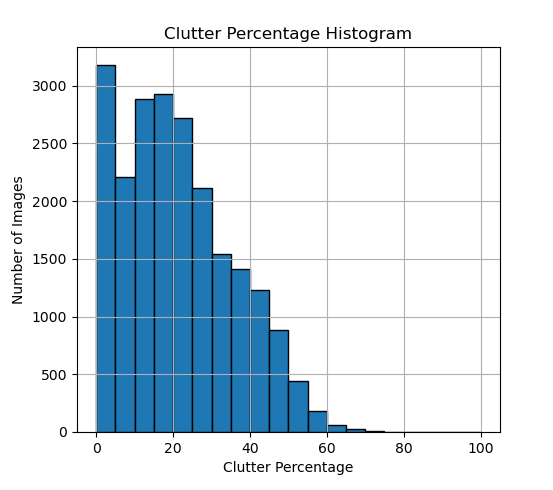}  
    \caption{\textbf{Clutter Distribution in the Structured3D dataset.} The histogram illustrates the number of scenes per 5\% interval of clutter.}
    \label{fig:clutter_stats}
\end{figure}

Figure \ref{fig:clutter_stats} illustrates the distribution of clutter across scenes, showcasing the number of scenes within 5\% clutter intervals. The majority of scenes exhibit a low percentage of clutter, with a mean clutter ratio of 21\% and the 75th percentile reaching 31\%.

\begin{figure*}[th!]
    \centering  
    \includegraphics[width=1.\linewidth]{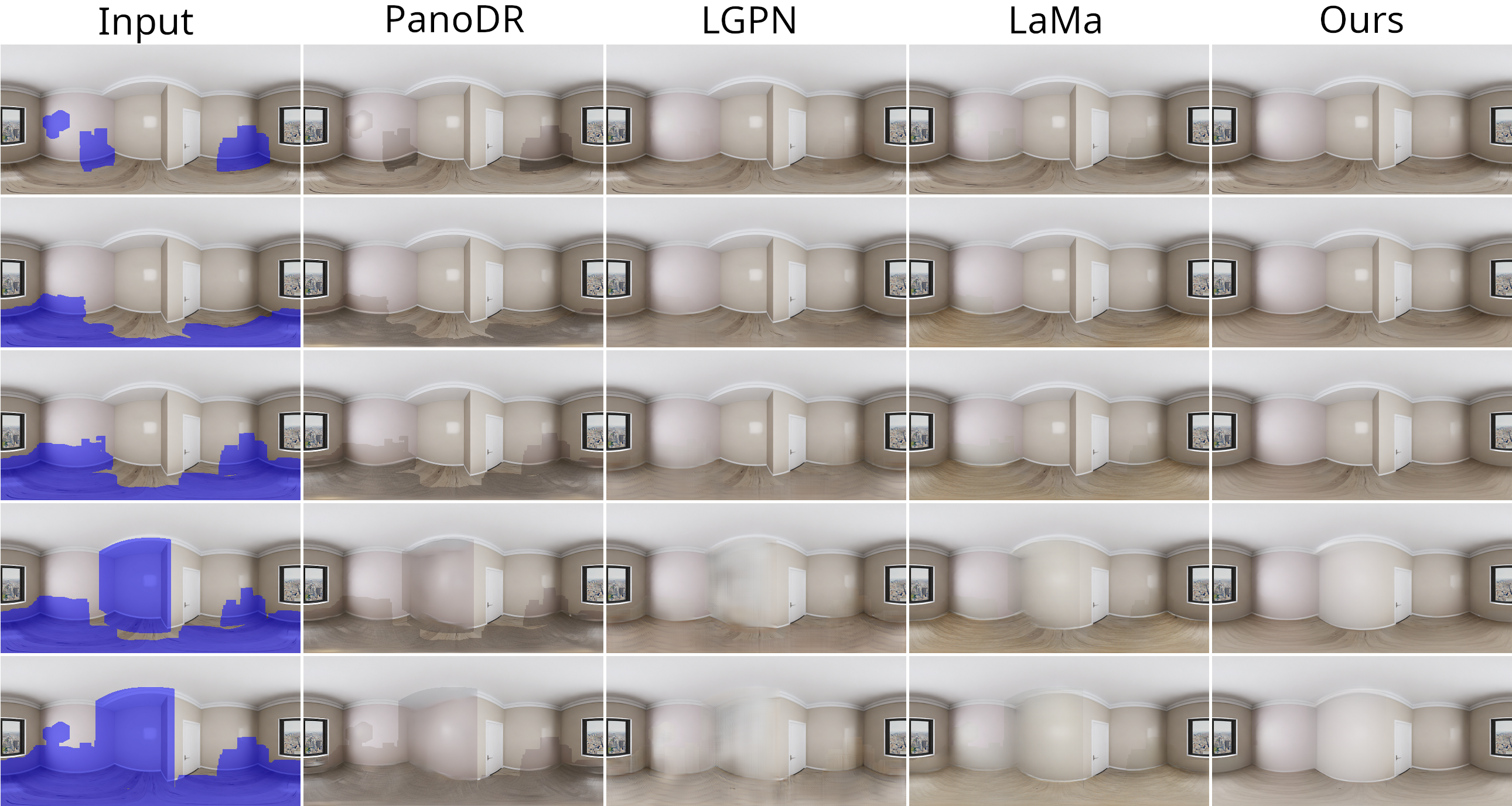}  
    \caption{\textbf{Qualitative results for the same room with different ratios of clutter .} The first row represents the 1 to 10 \% interval, then 10 to 20, 20 to 30, 30 to 40 and finnaly 40 to 50.}
    \label{fig:qualitative_comparison_clutter_ratio}
\end{figure*}
Figure \ref{fig:qualitative_comparison_clutter_ratio} illustrates the same room with different amounts of clutter, showing an example for each test interval. As we can see, the bigger the clutter amount, the more the floor is occluded, consequently, model must rely on ceiling line for reconstruction, showing the interest of symmetry property of the fourier unit.

\section{Inference time-memory comparison}
\label{sec:inference}
\paragraph{}
In this section, we compare different methods in terms of model size, inference time, and memory requirements. Results are reported using the provided code for PanoDR, LGPN, LaMa, and our implementation for our method, with a batch size of 1 averaged across 100 runs.
As depicted in Table \ref{tab:inference}, our model has a relatively large parameter count. This high number is attributed to the heavy use of gated convolution, which requires twice the convolutions (once for feature extraction and once for gating). We argue that this number can be reduced by employing lightweight gated convolutions with depthwise convolutions and replacing fusion learning (concatenation) in the W-FourierMixer with additive learning. Nevertheless, our method is faster and requires less memory than PanoDR and LGPN, being surpassed only by LaMa.
\begin{table}[h!]
    \centering
    \begin{tabular}{l|lll}
           & \#params\(\times10^6\) & inference time & memory    \\ \toprule
    Ours   &    104      &   77 ms             &   1.8GB        \\
    LaMa   &    27      &    4ms            &      0.8GB     \\
    PanoDR &    6      &     157ms           &      4.1GB     \\
    LGPN   &    124      &    504ms            &     5.8GB   
    \end{tabular}%
    \caption{}
    \label{tab:inference}
\end{table}

\section{Room Modeling}
\label{sec:room_modeling}
\begin{figure*}[t]
    \centering  
    \includegraphics[width=1.0\linewidth]{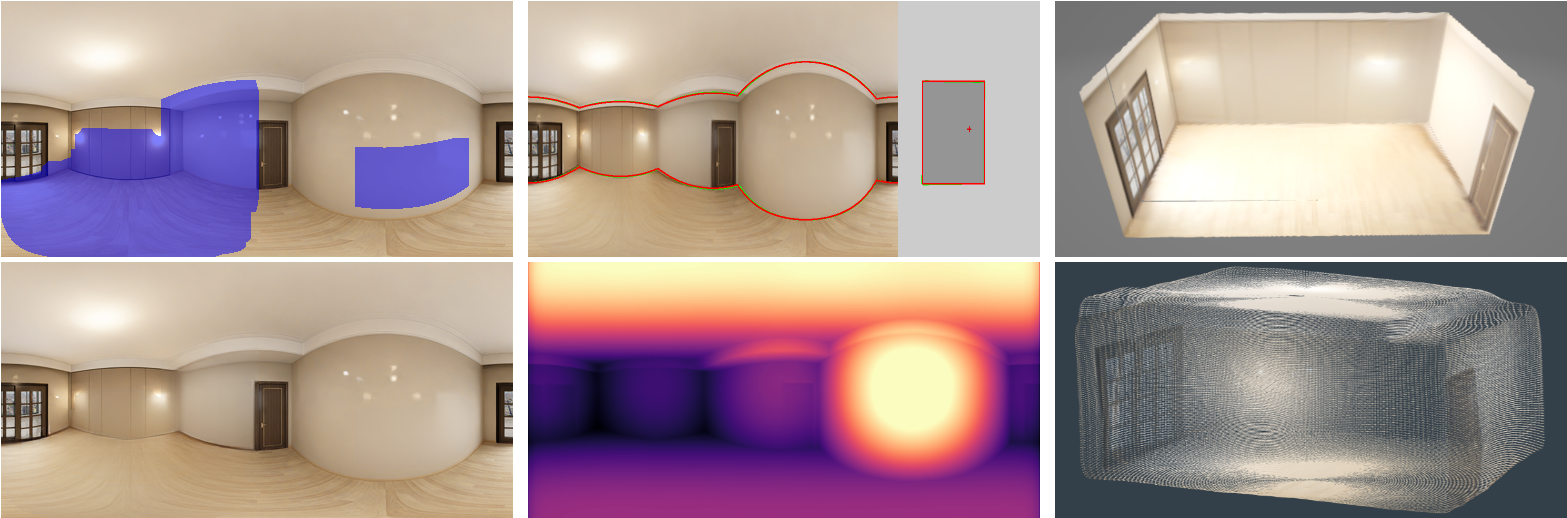}  
    \caption{\textbf{Results of inpainted room on RoomLayout and PanoDepth estimators.} First column represents input and inpainted prediction of our model, followed by the prediction of LGT-Net \cite{jiang2022lgt} (a room layout estimator) and the corresponding 3D reconstruction on the first row; the prediction of Joint\_360depth \cite{yun2022improving} network (a depth estimator) and the corresponding point cloud.}
    \label{fig:other_methods}
\end{figure*}
\paragraph{}
Our work is focused on removing clutter from indoor scenes to enhance room modeling, with a specific emphasis on creating textured, clutter-free environments. We intentionally omit 3D reconstruction in our method, as our primary goal is to produce structurally cohesive inpaintings. This intentional design allows our method to serve as a versatile preprocessing step for other techniques, see Figure \ref{fig:other_methods}, such as room layout estimators and depth estimation, facilitating subsequent 3D reconstruction processes.

\section{Additional qualitative results}
\label{sec:examples}
\begin{figure*}[t]
    \centering  
    \includegraphics[width=1.\linewidth]{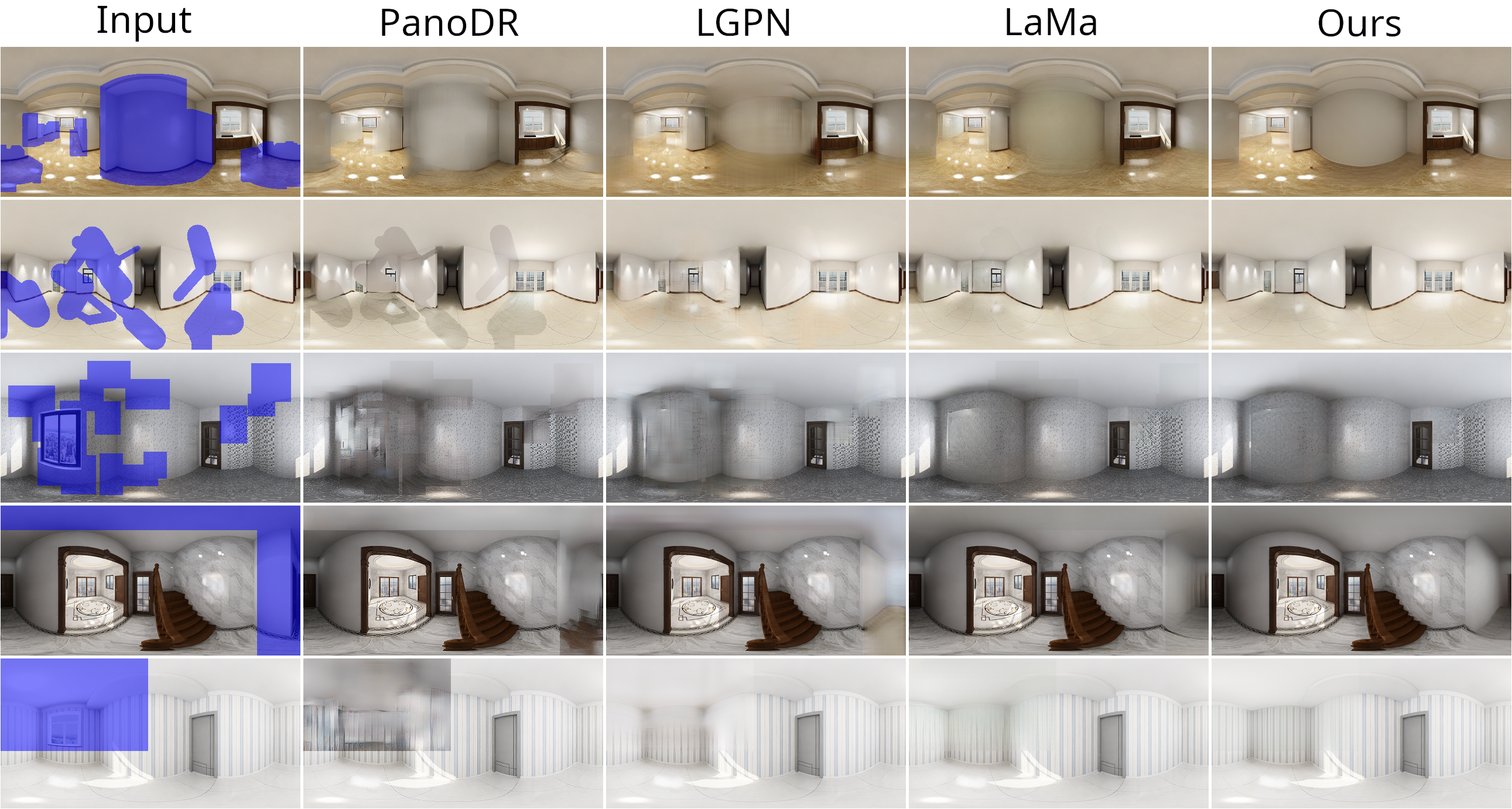}  
    \caption{\textbf{More qualitative results for different kinds of masks at arbitrary ratios.}}
    \label{fig:qualitative_comparison_supp}
\end{figure*}

%
%
%

\end{document}